\documentclass[pdflatex,sn-mathphys-num]{sn-jnl}

\usepackage{graphicx}%
\usepackage{multirow}%
\usepackage{amsmath,amssymb,amsfonts}%
\usepackage{amsthm}%
\usepackage{mathrsfs}%
\usepackage[title]{appendix}%
\usepackage{xcolor}%
\usepackage{textcomp}%
\usepackage{manyfoot}%
\usepackage{booktabs}%
\usepackage{algorithm}%
\usepackage{algorithmic}
\usepackage{listings}%
\usepackage{natbib}
\usepackage{subfigure}
\usepackage{multirow}

\theoremstyle{thmstyleone}%
\theoremstyle{thmstyletwo}%

\theoremstyle{thmstylethree}%

\begin{document}

\title[Article Title]{Fast Online Learning with Gaussian Prior-Driven Hierarchical Unimodal Thompson Sampling}


\author*[1]{\fnm{Tianchi} \sur{Zhao}}\email{zhaotianchi@bucea.edu.cn}
\equalcont{These authors contributed equally to this work.}

\author[2]{\fnm{He} \sur{Liu}}\email{liuhe@sem.tsinghua.edu.cn}
\equalcont{These authors contributed equally to this work.}

\author[1]{\fnm{Hongyin} \sur{Shi}}\email{shihongyin@bucea.edu.cn}

\author[2]{\fnm{Jinliang} \sur{Li}}\email{jll@tsinghua.edu.cn}

\affil*[1]{\orgdiv{School of Intelligence Science and Technology}, \orgname{Beijing University Of Civil Engineering And Architecture}, \orgaddress{\city{Beijing}, \postcode{102616}, \country{China}}}

\affil[2]{\orgdiv{School of Economics and Management}, \orgname{Tsinghua University}, \orgaddress{\street{Street}, \city{Beijing}, \postcode{100084},  \country{China}}}


\abstract{We study a type of Multi-Armed Bandit (MAB) problems in which arms with a Gaussian reward feedback are clustered. Such an arm setting finds applications in many real-world problems, for example, mmWave communications and portfolio management with risky assets, as a result of the universality of the Gaussian distribution. Based on the Thompson Sampling algorithm with Gaussian prior (TSG) algorithm for the selection of the optimal arm, we propose our Thompson Sampling with Clustered arms under Gaussian prior (TSCG) specific to the 2-level hierarchical structure. We prove that by utilizing the 2-level structure, we can achieve a lower regret bound than we do with ordinary TSG. In addition, when the reward is Unimodal, we can reach an even lower bound on the regret by our Unimodal Thompson Sampling algorithm with Clustered Arms under Gaussian prior (UTSCG). Each of our proposed algorithms are accompanied by theoretical evaluation of the upper regret bound, and our numerical experiments confirm the advantage of our proposed algorithms. 
}




\maketitle

\section{Introduction}
{The Gaussian distribution is widely observed in many physical and socioeconomic systems. Given two options whose reward follows Gaussian distributions, the ordinary way to compare them is to measure the reward of individual arms repetitively, and conclude the better one by running statistical tests with the two samples. }
\par 
{We explore a type of policy optimization in which the player or agent has the prior knowledge that the reward of an option follows a Gaussian distribution with mean and variance unknown. Further, we work with a unique optimal arm, the property \textit{Unimodality} which is expanded on in the next section. We present two applications as such in the following two examples.}

\begin{figure}[htbp!]
    \centering
    \includegraphics[width=0.77\linewidth]{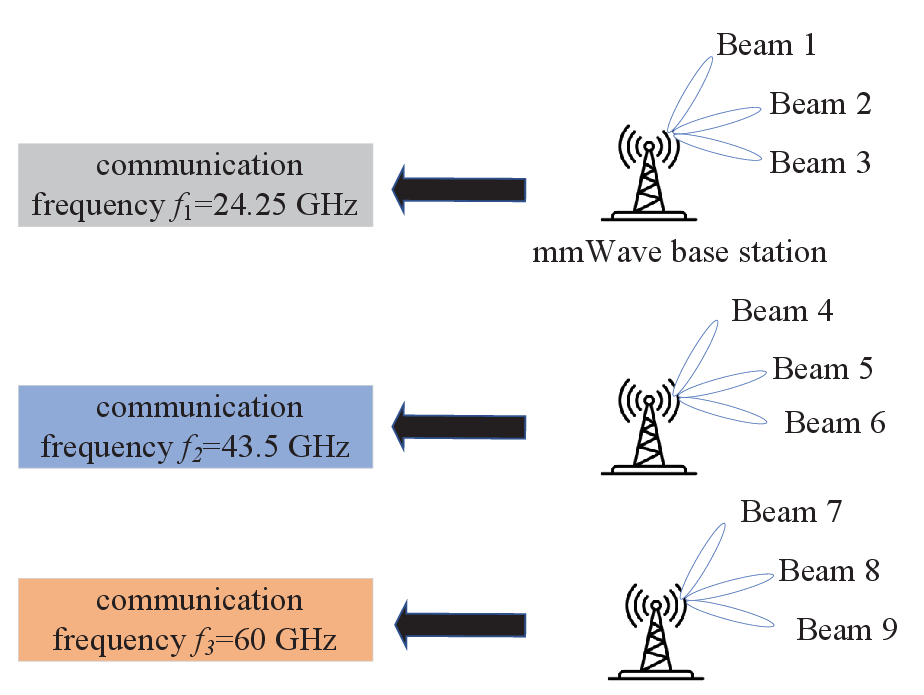}
    \caption{Example of mmWave communications. There are three communication frequencies: $f_1=24.25$ GHz, $f_2=43.5$ GHz, and $f_3=60$ GHz. For each communication frequency, there are three beams to be selected. }
    \label{fig:mmwave}
\end{figure}

\textbf{Example 1: mmWave communications}
\\
Millimeter wave (mmWave) technology is a key enabler in 5G communication, offering abundant bandwidth to accommodate the growing demand for mobile data traffic. However, a significant challenge in mmWave communication is its heavy reliance on the line-of-sight (LOS) path, which is highly susceptible to blockages from obstacles such as buildings. Beamforming plays a critical role in mmWave communications by enabling directional transmission that focuses the wireless signal toward the receiver. The choice of communication frequency directly affects large-scale path loss, while the beam selection determines the antenna gain. Thus, optimizing the selection of both the communication frequency and  beam  is essential for efficient mmWave communications. Furthermore, the received noise power typically follows a Gaussian distribution in mmWave communications. The objective of the learning agent (i.e., mmWave base station) is to identify the optimal communication frequency and beam that maximize the received signal strength in the presence of Gaussian noise. 
Fig. \ref{fig:mmwave} illustrates an example of mmWave communications. There are three communication frequencies \cite{3gpp_nr}: $f_1=24.25$ GHz, $f_2=43.5$ GHz, and $f_3=60$ GHz. For each communication frequency, there are three beams to be selected. In this context, each communication frequency can be considered as a cluster, with each  beam  representing an arm within that cluster.

\begin{figure}[h!]
    \centering
    \includegraphics[width=0.77\linewidth]{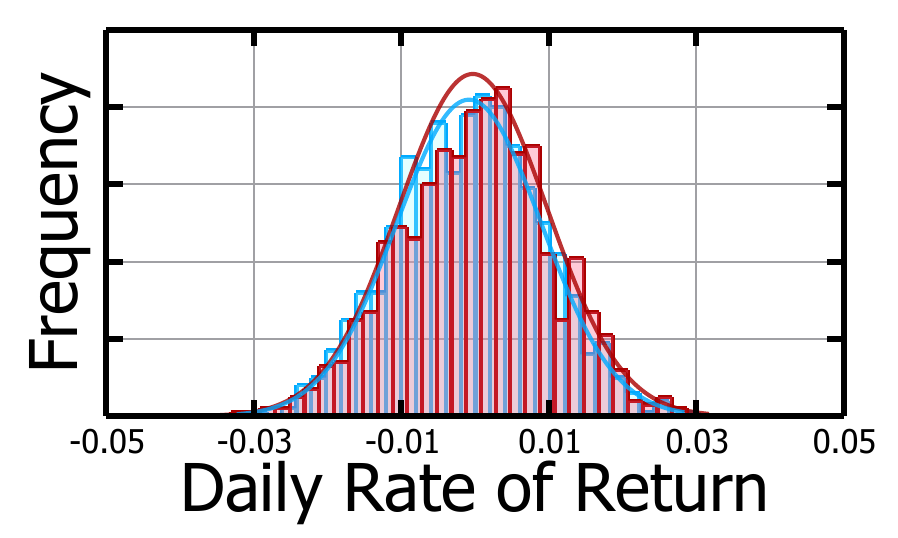}
    \caption{Example of Gaussian returns. The histogram in blue is synthesized according to statistics in S\&P 500 index in \protect\cite{Gaussian_SPX} and the one in red refers to CSI 300 statistics in \protect\cite{GaussReturns}.}
    \label{fig:Gauss_exp}
\end{figure}

\textbf{Example 2: Portfolio Selection}
\\
In short time periods, the return of a risky asset, such as one particular stock or a weighted allocation of many as a portfolio, is stochastic. In many quantitative models, the rate of return is approximated to by a superposition of a long-term drift and a Wiener process $\mathrm{d}W_t$ that accounts for Gaussian fluctuations.\footnote{We go ahead with this simple value process in this paper, as our primary concern is not the accurate description of financial time series.} An example is shown in Fig~\ref{fig:Gauss_exp}. 
\par

We start with a simple case, in which there are $n$ risky strategies as arms to choose from. Our aim is to find an automatic sampling scheme that gradually arrives at the optimal trading strategy which has the highest rate of return. Optimizations with both the mean and the variance in the Markowitz fashion can be found in \cite{MV3,MV1,MV2}, in which the agent maximizes the linear combination of the overall mean return and the risk measured by the variance, at a specific level of risk-tolerance.   
\par 
In each trading period or time slot, the AI-investor picks one asset and observes a return drawn from a Gaussian distribution that is specific to the asset and unknown to the investor. One practice as such can be found at \cite{shen2015portfolio}, which samples the arms in a UCB fashion. In our work, we examine selection in a more generalized setting, in which assets are clustered. 
\begin{table}[h!]
    \centering
    \begin{tabular}{cccc}
    \hline
    stock &market&return&std $\sigma$\\
    \hline
        CSI\#1 & China & 0.00060& 0.010\\
        CSI\#2 & China & 0.00062& 0.010\\
        CSI\#3 & China & 0.00058& 0.010\\
        SPX\#1 & U.S. & 0.00040& 0.010\\
        SPX\#2 & U.S. & 0.00042& 0.010\\
        SPX\#2 & U.S. & 0.00037& 0.010\\
        \hline
    \end{tabular}
    \caption{Sample return statistics of different risky assets. Our construction of this demonstration refers to the statistics in \protect\cite{GaussReturns}.}
    \label{tab:my_label}
\end{table}
\par 
The hierarchy and Unimodality of arms in systems above can be dealt with algorithms specific to utilize the two properties. We first review previous work {that does such} later in this section, and define our setup in Section.~\ref{S2}. We introduce TSG in Section.~\ref{sec:TSG} as the base algorithm, upon which we propose our first TSCG in Section.~\ref{sec:TSCG} and UTSCG in Section.~\ref{sec:UTSCG}. The two algorithms are tested with experiments in Section.~\ref{sec:exp}. Section.~\ref{sec:discussion} concludes.
\subsection{Related Work}
\subsubsection{Bandits with hierarchical structures}
Extensive research has been conducted on the hierarchical multi-armed bandit (MAB) problem, where the set of available arms is partitioned into several distinct clusters \cite{nguyen2014dynamic,jedor2019categorized,bouneffouf2019optimal,carlsson2021thompson}. Under different premises regarding clustering configurations, these existing studies have derived corresponding regret bounds for their proposed methods. For instance, a Two-level Policy (TLP) algorithm, which groups arms into multiple clusters, was first introduced in \cite{pandey2007multi}. A limitation of this work is the absence of theoretical assessments regarding the algorithm’s upper bound on regret.
\cite{zhao2019fast} proposed a novel Hierarchical Thompson Sampling (HTS) algorithm. In a relevant context, the beams associated with a single selected group can be analogously treated as a cluster of arms within the MAB framework. Even so, this HTS approach fails to leverage the Unimodal characteristic inherent to each individual cluster. Another notable contribution is the two-level UCB mechanism proposed in \cite{bouneffouf2019optimal}, which operates under the assumption that the arm set is pre-clustered and that arms within the same cluster exhibit similar reward distributions. Like the HTS algorithm, this UCB-based method does not account for the Unimodal property of each cluster. Meanwhile, \cite{jedor2019categorized} described a specific MAB scenario where arms are classified into three distinct types, with each type featuring a unique inter-cluster ordering. In \cite{yang2024optimal}, the focus shifts to an online clustering problem, wherein a collection of arms can be divided into various unknown groups. It is important to note that the clustering partition in this work is time-varying, which differs fundamentally from our research setting where clusters are pre-defined and fixed. The problem of hidden population sampling in online social platforms was addressed in \cite{kumar2019hierarchical}, where a hierarchical MAB algorithm named Decision-Tree Thompson Sampling (DT-TMP) was proposed. This algorithm integrates a decision tree with a reinforcement learning search strategy to explore the combinatorial search space, though it lacks theoretical analysis of its performance. A different line of research is presented in \cite{zhao2020regret,singh2024multi}, which examines MAB problems involving dependent arms. In their setting, pulling arm $i$ not only uncovers information about the reward distribution of arm $i$ itself but also reveals relevant information about all other arms in the same cluster as arm $i$—a scenario that does not apply to our problem. Additionally, \cite{sen2021top} adopted a specific sampling strategy within a hierarchical MAB-based top-k contextual recommendation model and validated the resulting reduction in regret through theoretical proofs. For clustered arms, \cite{carlsson2021thompson} developed a Thompson Sampling-based algorithm (TSC) and derived a regret bound that is dependent on the number of clusters, yet this method also does not make use of the Unimodal property. A unified framework for multitask learning based on hierarchical MAB was proposed in \cite{hong2022hierarchical}, accompanied by a comprehensive regret analysis. Furthermore, the Unimodal Two-Level Policy (UTLP) algorithm was introduced in \cite{Zhao_Zong_Zhang_2026}, which similarly lacks rigorous theoretical analysis. Beyond this, \cite{ZhaoLiuLiuLiZhaoShiLi_2025} presented the Modified Unimodal Thompson Sampling Algorithm with Clustered Arms (MUTSC), a method that focuses on Beta priors and is therefore incompatible with our current research problem.
 \subsubsection{Unimodal Bandit} A variety of specialized algorithms have been developed for Unimodal bandit problems, each incorporating distinct exploration mechanisms—representative examples include the Kullback-Leibler Upper Confidence Bound (KL-UCB) and Thompson Sampling (TS) methods. The pioneering work in \cite{yu2011Unimodal} laid the foundation for investigating this problem under both continuous and discrete arm configurations. Building on this, \cite{combes2014Unimodal} proposed the Optimal Sampling for Unimodal Bandits (OSUB) algorithm, which makes full use of the Unimodal structure across both continuous and discrete arm settings. For this OSUB algorithm, the research team derived an upper bound on regret that is not affected by the number of arms. Subsequent research by \cite{paladino2017unimodal} advanced this field by designing a Thompson Sampling-based method specifically for Unimodal scenarios, termed the Unimodal Thompson Sampling (UTS) algorithm. Drawing on the findings presented in \cite{paladino2017unimodal}, \cite{trinh2020solving} delivered a rigorous and precise theoretical analysis of the UTS algorithm. Later, \cite{zhang2021mmwave} adapted the analytical framework proposed by \cite{trinh2020solving}, extending the proof scope from Bernoulli arms to multinomial arms. In the context of mmWave links, \cite{hashemi2018efficient} developed a solution based on the MAB framework for beam alignment tasks. This method leverages the Unimodal characteristic of the average received signal strength (RSS) to exploit the inherent correlation of channels; by excluding beams with subpar performance, it effectively narrows down the search space to the region surrounding the optimal beam. In \cite{zhao2022hierarchical}, the research focus turned to bandit problems with clustered arms, where the expected reward within each cluster follows a Unimodal distribution. This theoretical framework is applicable to practical scenarios such as multi-channel mmWave beam selection and codebook selection, and their results inherently depend on the number of arms in each cluster.
 Most recently, \cite{yang2023thompson} established a regret bound for the Thompson Sampling algorithm with a Gaussian prior in the context of Unimodal Bandits under Gaussian reward distributions. Their theoretical proofs demonstrate that the asymptotic regret of the proposed algorithm achieves $O(\log(T))$, which matches the regret order of existing UTS algorithms.
\subsection{Main Contributions}
Our main contributions are summarized as follows:
\begin{enumerate}
\item We identify a family of optimization problems, in which available options are clustered with individual Gaussian feedback, and the optimal option is unique. We present a systematic description of how we can convert the optimization to a Unimodal-hierarchical bandit learning task. In a MAB fashion, we are able to design efficient sampling algorithms that utilize arm hierarchy and Unimodality.   
\item We propose two improved algorithms based on the TSG algorithm. Specifically, first we propose a TSCG algorithm to accommodate and utilize the structure of clusters. Through theoretical analysis, we show that in TSCG we can achieve a smaller upper bound on the regret compared to the TSG algorithm, due to the improved utilization of cluster structure properties. Second, we propose a UTSCG algorithm, which leverages the Unimodal property. Our analysis reveals that the upper bound on the regret in UTSCG is even lower than that of the TSCG algorithm, demonstrating the additional benefits of incorporating the Unimodal property into the algorithm design. 
\item Our two proposed algorithms  are verified by experiments in different arm configurations. The proposed algorithms outperform baseline algorithms, TSG,  UCB and TLP. We validate our algorithms' efficiency and effectiveness in terms of the cumulative regret and the percentage of optimal arm selected over time. The improvements in the upper bound of the regret by theory are all confirmed by the numerical results.
\end{enumerate}

\section{System Model}
\label{S2}
\subsection{Gaussian Arms}
We study a system of $n$-arms that are clustered to $K$ subsets $\mathcal{K}=\{C_1,C_2,...,C_K\}$, which forms a partition of the index set $\left[n \right] = \{1,2,...,n\}$. The time horizon spans $T$ rounds and in each round, the agent picks one arm and receives a stochastic reward $X_i(t)$ of arm $i$ that is drawn from a Gaussian distribution
\begin{equation}
   X_{i}(t) \sim N(\mu_i,\sigma_i^2)
\end{equation}
where $\mu_i$ and $\sigma_i$ are specific to the arm and constant throughout the time window. In most real world scenarios, the agent does not have the knowledge of the mean $\mu$.
\subsection{Regret Minimization}
In this work, we examine the case in which the objective is to maximize the cumulative reward, phrased by the optimization:
\begin{align}
    \max_{\{i(1),i(2),...,i(T)\}} \sum_{t=1}^{T}X_{i(t)}(t).
\end{align}
The maximization is converted to the minimization of the expected \textit{regret}, the loss w.r.t. the expected total reward as if the optimal arms were selected the whole time:
\begin{equation}
    E\left[R(T)\right] = T \mu^* - E\left[\sum_{t=1}^{T} X_{i(t)}(t)\right],
\end{equation}
where  $\mu^*$ refers to the expectation of the optimal arm $i^*$
\begin{equation}
    \mu^* = \max_{i} \left[\mu_i\right]
\end{equation}
Note that there is a different route that only optimizes the maximal one-time reward witnessed up till the terminal time slot $t=T$ \cite{pure_explore}.
\subsection{Arm configurations}
In the two examples, the expected reward of each arm depends on two attributes. In mmWave communications, arms are specified by the frequency and spatial orientation according to the theory of {wireless communication}; in portfolio selection, the property of an arm depends on the market it belongs {to}, and the construction of the portfolio, for example the weighting of the risky free bond. To describe this structure, we define a measure for $\mu$ parametrized by two variables $x^1$ and $x^2$. Arms are clustered by $x^1$ which can be continuous or categorical, and one label arms according to the ascending or the descending order of $x^2$ in each cluster. We highlight that our two proposed algorithms achieve a ``cluster-invariance'', i.e., the expected upper bound on the regret is unchanged under a swapping  $C_i \leftrightarrow C_j, \,\forall i,j$.
\subsubsection{Unimodality}
Our algorithms are designed for hierarchical arms that are Unimodal. Arms are evaluated by the expected reward, in terms of parameter $\mu$. We require that:
\begin{enumerate}
    \item the optimal arm $i^*$ is unique;
    \item in each cluster, the expected reward has a unique maximal, and the change in $\mu$ w.r.t. the index is monotonic post or prior to $i^*$.
\end{enumerate}

\begin{figure}[h!]
\centering~
\subfigure{
\label{uni_peak}
\includegraphics[width=0.25\textwidth]{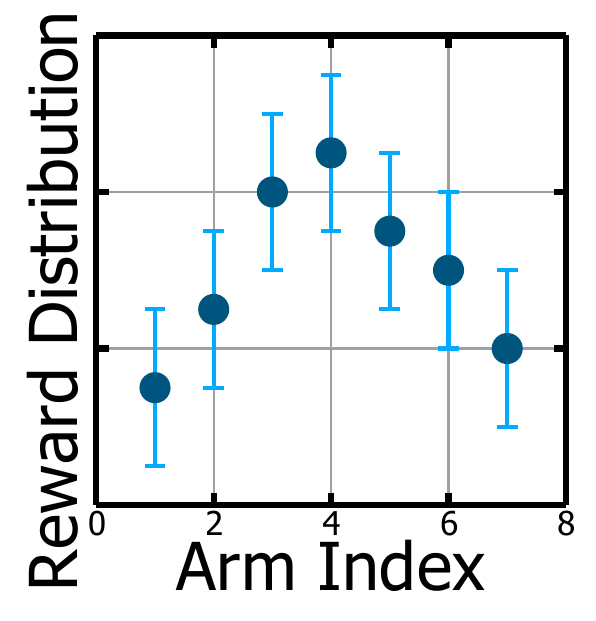}}
\subfigure{
\label{uni_decrease}
\includegraphics[width=0.25\textwidth]{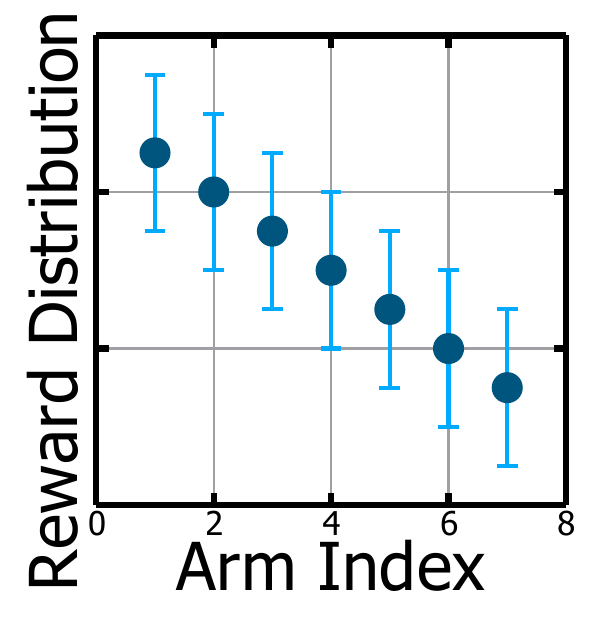}}
\subfigure{
\label{uni_increase}
\includegraphics[width=0.25\textwidth]{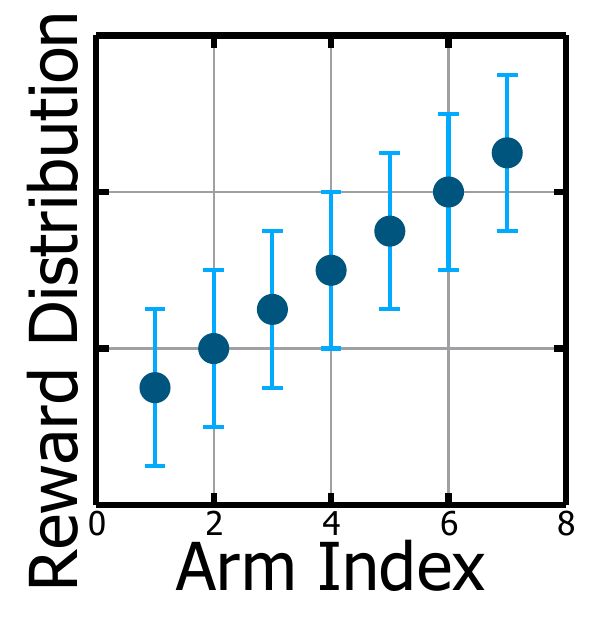}}
\caption{Three Unimodal configurations are presented. The vertical bar indicates the standard deviation. There can be either one unique maximum in the expected regret, or the mean monotonically decreases or increases with the arm labeling in the cluster.}
\label{fig:hier_unimodal}
\end{figure}
To demonstrate Unimodality inside the cluster, we show in Fig.~\ref{fig:hier_unimodal} three possible configurations. Note that the optimal arm has different number of neighboring arms. When $\mu$ is monotonic in a cluster, the optimal arm has only 1 neighbor, while there are 2 neighbors when the pattern is a typical peak. 
\section{Thompson Sampling with Gaussian Prior}
\label{sec:TSG}
We begin by introducing Thompson Sampling with a Gaussian prior (TSG) as the foundation for our subsequent algorithms. We present Theorem 1 to demonstrate the regret bound of TSG. The problem-dependent regret bound provided by \cite{yang2023thompson}  
{poses challenges for application in the analysis of our subsequently proposed algorithms due to the difficulty in leveraging the hierarchical property.}
The basic idea of TSG can be summarized as follows: the algorithm assumes that the reward of each arm follows a Gaussian distribution (with varying means). In each round, the decision-maker first samples the reward for each arm  from its prior distribution. The arm with the highest sampled reward is then selected, and the corresponding posterior is updated with the reward $X_i(t)\sim N(\mu_i,1)$. Typically, the normal distribution $N(\hat{\mu}_i(t-1),\frac{1}{N_i(t-1)+1})$ is used as the prior for arm $i$ in round $t$. It is straightforward to compute the posterior distribution as a Gaussian distribution $N(\hat{\mu}_i(t),\frac{1}{N_i(t)+1})$, where $\hat{\mu}_i(t)$ and $N_i(t)$ are updated based on the observed rewards. The TSG’s process is detailed in Alg.~\ref{alg:TSG}.
\begin{algorithm}
\caption{Thompson Sampling Algorithm with Gaussian Prior (TSG)}  
\label{alg:TSG}  
\begin{algorithmic}[1]
\STATE {$\forall i$, set $\hat{\mu}_i(0)=0,N_i(0)=0$}  
\FOR{ $t=1,2,...,T$}
\STATE For each arm $i=1,\ldots, n$, sample $\eta_i(t)$ from the $N(\hat{\mu}_i(t-1),\frac{1}{N_i(t-1)+1})$ distribution.
\STATE Play arm $i(t):=\arg\max_i \eta_i(t)$ and observe Gaussian reward $X(t)$.
\STATE Update empirical mean $\hat{\mu}_{i(t)}(t)=\frac{\hat{\mu}_{i(t)}(t-1)\times N_{i(t)}(t-1)+X(t)}{N_{i(t)}(t-1)+1}$, $N_{i(t)}(t)=N_{i(t)}(t-1)+1$.
\ENDFOR
\end{algorithmic}  
\end{algorithm}
 



First, we prove the following lemma in order to derive Theorem 1, as Lemma 3 in \cite{Agrawal2012FurtherThompson} does not provide an extension result for the Gaussian distribution.

\textbf{Lemma 1}  Let $X_i(1)...X_i(T)$ be i.i.d  $N(\mu_i,\sigma^2)$, $\hat{\mu}_i(T)=\frac{1}{T}\sum_{t=1}^{T}X_i(t)$. Then,
\begin{eqnarray}
\label{eq:lemma1_1}
\sum_{t=1}^{T}P(i(t)=i,\hat{\mu}_i(t)\geq x_i)\leq 1+\frac{2e^{-\frac{\Delta_i^2}{18\delta^2}}}{1-e^{-\frac{\Delta_i^2}{18\delta^2}}}.  
\end{eqnarray}
where $\Delta_i=\mu_{i^*}-\mu_i$, $x_i$ is a threshold which satisfies $x_i=\mu_i+\frac{\Delta_i}{3}$.

\textbf{Proof outline for Lemma 1}: This result mainly follows the sub-Gaussian Chernoff-Hoeffding bounds, which provide insights to the concentration of the estimation $\hat{\mu}_i(T)$. For more details, please refer to the supplementary material.

\textbf{Theorem 1} \label{thm:1}For an $n$-armed stochastic bandit problem, TSG has expected regret
\begin{eqnarray}
\label{eq:Thm1_2}
E[R(T)]\leq\sum_{i=1}^{n}\left(\frac{18\log(D_{max}T)}{\Delta_i}\right)+O(n).    
\end{eqnarray}
Where $D_{max}=\max_{i}\Delta_i^2$.

\textbf{Proof outline for Theorem 1}: The analysis of the TSG follows the proof procedure of the TS in \cite{Agrawal2012FurtherThompson}. Compared to the TS algorithm, we highlight key differences in the proof: 
        \begin{enumerate}
        \item To adapt our analysis to the Gaussian distribution, we apply the result of Lemma 1 instead of Lemma 3 from \cite{Agrawal2012FurtherThompson}. 
        \item We did not apply the proof procedure outlined in \cite{Agrawal2012FurtherThompson} to derive a problem-independent bound for $L_i(T)=\frac{18\log(T\Delta_i^2)}{\Delta_i^2}$.
    \end{enumerate}
    Due to the space limitation, we put the complete proof in the supplementary material.
\section{Thompson Sampling with Clustered Arms under Gaussian Prior (TSCG)}
\label{sec:TSCG}
Given a clustering of the arms into a set of clusters $\mathcal{K}$, we introduce the Thompson Sampling algorithm with Clustered Arms under Gaussian Prior (TSCG), Alg.~\ref{alg:TSC}. In addition to maintaining a belief for each arm $i$, represented as $N(\hat{\mu}_i(t),\frac{1}{N_i(t)+1})$, the agent also keeps a belief over possible expected rewards $N(\hat{\mu}_C(t),\frac{1}{N_C(t)+1})$ for each cluster $C\in \mathcal{K}$. At each time slot $t$, the agent first uses TS to select a cluster. This is done by sampling $\eta_C(t)\sim$ $N(\hat{\mu}_C(t-1),\frac{1}{N_C(t-1)+1})$ for each cluster $C\in \mathcal{K}$ and then considering the cluster $C(t)=\arg\max_{C\in \mathcal{K}}\eta_C(t)$ (line 3). In lines 4 to 5, the agent samples $\eta_i(t)\sim N(\hat{\mu}_i(t-1),\frac{1}{N_i(t-1)+1})$ for each arm $i\in C(t)$ and plays the arm $i(t)$. Then the agent updates the belief for $i(t)$ and $C(t)$ as shown in line 6 in Alg.~\ref{alg:TSC}.

\begin{algorithm}[h!]
\caption{Thompson Sampling Algorithm with clustered arms under Gaussian Piror (TSCG) }  
\label{alg:TSC}  
\begin{algorithmic}[1]
\STATE {$\forall i$ and $\forall C$,set $\hat{\mu}_i(0)=0,N_i(0)=0, \hat{\mu}_C(0)=0,N_C(0)=0$}  
\FOR{ $t=1,2,...,T$}
\STATE For each cluster $C$, sample $\eta_C(t)$ from the $N(\hat{\mu}_C(t-1),\frac{1}{N_C(t-1)+1})$ distribution and pick $C(t)=\arg\max_{C\in \mathcal{K}}\eta_C(t)$.
\STATE For each arm $i\in C(t)$ sample  $\eta_i(t)$ from the $N(\hat{\mu}_i(t-1),\frac{1}{N_i(t-1)+1})$. 
\STATE Play arm $i(t)=\arg\max_i \eta_i(t)$ and observe Gaussian reward $X(t)$.
\STATE Update empirical mean $\hat{\mu}_{i(t)}(t)=\frac{\hat{\mu}_{i(t)}(t-1)\times N_{i(t)}(t-1)+X(t)}{N_{i(t)}(t-1)+1}$, $N_{i(t)}(t)=N_{i(t)}(t-1)+1$,$\hat{\mu}_{C(t)}(t)=\frac{\hat{\mu}_{C(t)}(t-1)\times N_{C(t)}(t-1)+X(t)}{N_{C(t)}(t-1)+1}$, $N_{C(t)}(t)=N_{C(t)}(t-1)+1$.
\ENDFOR
\end{algorithmic}  
\end{algorithm}

Let us assume that we have a clustering of $n$ arms, grouped into a set of clusters $\mathcal{K}$. For each arm $i$, the expected reward is denoted by $\mu_i$. Let $i^*$ be the unique optimal arm with expected reward of $\mu_{i^*}$. The cluster containing $i^*$ is denoted as $C^*$.  For each cluster $C\in \mathcal{K}$, we define $\overline{\mu}_C=\max_{i\in C}\mu_i$, $\underline{\mu}_C=\min_{i\in C}\mu_i$, $\Delta_C=\mu_{i^*}-\underline{\mu}_C$ and $\Delta_{C}'=\underline{\mu}_{C^*}-\overline{\mu}_C$. For each cluster $C\in \mathcal{K}$, we define the distance $d_C=\min_{i\in C^*,\hat{i}\in C}\mu_i-\mu_{\hat{i}}$ and the width $w_C=\overline{\mu}_C-\underline{\mu}_C$, let $w^*$ denote the width of the optimal cluster.

\textbf{Assumption 1} (Strong Dominance) For $C\neq C^*$, $d_C>0$.

This assumption implies that the expected value of each arm in the optimal cluster is better than the expected value of each arm in any sub-optimal cluster. In order to bound the regret of TSCG, we will bound the number of plays of a sub-optimal cluster $N_C(T)$ in the following lemma.

\textbf{Lemma 2} Under Assumption 1, the expected number of plays of a sub-optimal cluster $C$ at time $T$ using TSCG is bounded by
\begin{eqnarray}
\label{eq:Lemma2_5}
E[N_C(T)]&&\leq \frac{18\log(D_{max}T)}{\Delta_{C}'^2}+O(1).
\end{eqnarray}
\textbf{Proof outline for Lemma 2}: The regret analysis of our Lemma 2 mainly follows the streamline of proving Lemma 2 in \cite{carlsson2021thompson}. However, to accommodate arm clustering  with TSCG, we have made some modifications to the proof in \cite{carlsson2021thompson}. We highlight two key differences here: 
        \begin{enumerate}
        \item the treatment of the event that, one arm in cluster $C$ is played at time $t$,  follows the splitting used in \cite{Agrawal2012FurtherThompson} instead of the manner in \cite{carlsson2021thompson}.
        \item We refer to our own result, Theorem 1, as the upper bound of the number of plays of a sub-optimal cluster $C$, as our problem setting is Gaussian while Lemma 2 in \cite{carlsson2021thompson} deals with Bernoulli arms.
    \end{enumerate}

For full details of our proof, please refer to the supplementary material. With Lemma 2, we can derive the following instance-dependent regret bound for TSCG.

\textbf{Theorem 2} For an $n$-armed stochastic bandit problem under Assumption 1, the TSCG has an expected regret 
\begin{eqnarray}
\label{eq:proof8}
E[R(T)]&&\leq \sum_{C\neq C^*}\frac{18\log(D_{max}T)}{\Delta_{C}'^2}\Delta_{C}+o(\log(T))\nonumber
\\
&&+ \sum_{i\in C^*}\frac{18\log(D_{max}T)}{\Delta_i}+O\left( n\right).  
\end{eqnarray}

\textbf{Remark} Theorem 2 demonstrates that the upper bound of the expected regret depends on:
\begin{enumerate}
    \item the number of clusters, shown in the first summation;
    \item the number of arms in the optimal cluster, explicit in the second summation;
    \item the quality of the clustering, which is measured by $\Delta_{C}'$.
\end{enumerate}
Notice that $E(R(T))$ is independent of the total number of arms $n$.

\textbf{Proof outline for Theorem 2}: The analysis of the TSCG follows the proof procedure of the TSC in \cite{carlsson2021thompson}. Compared to the TSC algorithm, we highlight key differences in the proof: 
        \begin{enumerate}
        \item We use our Lemma 2's result, instead of that in \cite{carlsson2021thompson}, to bound the expected plays of sub-optimal clusters for Gaussian distributions.
        \item We apply the result of our Theorem 1 to bound the expected plays of sub-optimal arms, as our problem setting is Gaussian.
    \end{enumerate}
\section{Unimodal Thompson Sampling Algorithm with Clustered Arms under Gaussian Prior (UTSCG)}
\label{sec:UTSCG}
Both the TSG and TSCG algorithms do not fully exploit the Unimodal property within each cluster. In this section, we propose a new algorithm that utilizes the Unimodal property. Our proposed algorithm, which we refer to as the Unimodal Thompson Sampling Algorithm with Clustered Arms under Gaussian Prior (UTSCG), is built upon the Thompson Sampling for Unimodal Bandit with Gaussian Prior \cite{yang2023thompson} and TSCG. The UTSCG, detailed in Alg.~\ref{alg:UTSC}, offers a lower regret bound compared to the TSCG algorithm.
\begin{algorithm}[h!]
\caption{Unimodal Thompson Sampling Algorithm with cluster arm under Gaussian Prior (UTSCG)} 
\label{sec:UTSC}
\label{alg:UTSC}  
\begin{algorithmic}[1]
\STATE {$\forall i$ and $\forall C$, set $\hat{\mu}_i(0)=0,N_i(0)=0, \hat{\mu}_C(0)=0,N_C(0)=0$}  
\FOR{ $t=1,2,...,T$}
\STATE For each cluster $C$, sample $\eta_C(t)$ from the $N(\hat{\mu}_C(t-1),\frac{1}{N_C(t-1)+1})$ distribution and pick $C(t)=\arg\max_{C\in K}\eta_C(t)$
\STATE Find the leader $L(t)=\arg\max_{i\in C(t)}\hat{\mu}_{i}(t)$.
\STATE For each arm $i\in \{Neighbor(L(t))\cup L(t)\}$, sample $\eta_i(t)$ from the $N(\hat{\mu}_i(t-1),\frac{1}{N_i(t-1)+1})$ distribution.
\STATE Select arm $i$, where $i(t)=\arg\max_i \eta_i(t)$ and observe Gaussian reward $X(t)$
\STATE Same as line 6 of Alg.~\ref{alg:TSC} to update $\hat{\mu}_i(t),N_i(t), \hat{\mu}_C(t),N_C(t)$.
\ENDFOR
\end{algorithmic}  
\end{algorithm}

The algorithm operates as follows: at each time slot $t$,  the agent first uses TSG to select a cluster, as indicated in line 3.  After that, in line 4, UTSCG selects the arm, denoted as the leader $L(t)$ in round $t$, by determining the arm with the maximum empirical expected utility value. Once the leader is chosen, the selection process is confined to the leader $L(t)$ and its adjacent neighborhoods. We denote $\gamma_{L(t)}$ as the number of {neighboring arms next to} the leader $L(t)$. Then, from line 5 to line 6, the TSG algorithm is performed over arm $i$ such that $i\in \{Neighbor(L(t))\cup L(t)\}$, where $Neighbor(i)$ denotes the set of neighbors of arm $i$.

\textbf{Assumption 2} (Unimodality in each cluster) For $\forall C\in\mathcal{K}$, the expect reward function among the cluster $C$ is a Unimodal function.

\textbf{Theorem 3} For an $n$-armed stochastic bandit problem under Assumptions 1 and 2,  the UTSCG has an expected regret of 
\begin{eqnarray}
\label{eq:proof29}
&&E[R(T)]\leq\sum_{C\neq C^*}\frac{18\log(D_{max}T)}{\Delta_{C}'^2}\Delta_{C}+o(\log(T))\nonumber
\\
&&+ \sum_{i\in Neighbor(i^*)}\frac{18\log(D_{max}T)}{\Delta_i}+O\left(n\right)+F.  
\end{eqnarray}
{Where $F=\sum_{i\neq i^*}\left(\frac{12}{\Delta_i}+\frac{16\log(2\Delta_i)+4}{\Delta_i\sqrt{2\pi}}+2\sqrt{2\pi}\right)$ \cite{yang2023thompson}.} Compared with the TSCG algorithm, it is evident that the UTSCG algorithm exhibits a lower regret. This lower regret is primarily attributed to the effect of the third term in Theorem 3, which measures the regret incurred within the optimal cluster $C^*$. This term does not depend on the number of arms in the optimal cluster $C^*$  (in contrast to the TSCG algorithm, which may have a regret bound that scales with the number of arms in the optimal cluster).

\textbf{Proof outline for Theorem 3}: The analysis of the UTSCG follows the proof procedure of the TSCG. Compared with the TSCG algorithm, we highlight key differences in the proof: for each cluster $C$ (including $C^*$), we break down the regret incurred by selecting sub-optimal arms in each cluster into two parts:
    \begin{enumerate}
        \item The regret resulting from the algorithm's selection of the neighbor of the best arm $i_C^*$ within  cluster $C$ is analyzed by referring to the result stated in Theorem 1.
        \item The regret arising from the algorithm's selection of a non-neighbor of the best arm $i_C^*$ in cluster $C$ is analyzed by applying the result from step 3 in Ref.~\cite{yang2023thompson}.
    \end{enumerate}
    Due to the space limitation, we omit the proof here (proved in the supplementary material).
\section{Numerical Experiments}
\label{sec:exp}
We test our TSCG and UTSCG in two scenarios that we have described in the first section. The two proposed algorithms are evaluated by the cumulative regret as well as the cumulative rate of the true optimal arm having been selected.
\subsection{mmWave Communications}
The mmWave communications are simulated using MATLAB. We choose three communication frequencies from the 3rd Generation Partnership Project (3GPP) specification \cite{3gpp_nr}:  $f_1=24.25$ GHz, $f_2=43.5$ GHz, and $f_3=60$ GHz. We regard the beams under the same communication frequency as a cluster and each beam as an arm. For each communication frequency (cluster), there are a total of 3 beams. For the wireless channel, we choose the 3GPP standard statistic free-space signal propagation model in mmWave communications \cite{sun2017smart}. The large-scale path loss is modeled as:  
\begin{equation}
PL(d)[dB]= 20 \log_{10}(f) + 20 \log_{10}(d) + 92.45,
\end{equation}
where $d$ is the distance between a transmitter (Tx) and a receiver (Rx) in kilometers, $f$ is the communication frequency in GHz. 

The received signal strength (RSS) at a Rx is  
\begin{equation}
\label{rss}
RSS = P_{Tx} \times G \times PL(d)^{-1} + N,  
\end{equation}
where $P_{Tx}$ is the transmission power of Tx, $G$ is the antenna gain. The antenna gain in mmWave communications highly depends on the direction of beams formed by the Tx and Rx, $N$ is the Gaussian noise in Rx.   We set the transmission power $P_{Tx}$ as 60 dBm, and 
\begin{equation} 
    G=\left\{
\begin{aligned}
& G_{max}, & \text{if} \ |\omega| < \omega_s\\
& G_{min}, & \text{otherwise,}\\
\end{aligned}
\right.
\end{equation}
where $G_{max}$ is the main lobe gain, $G_{min}$ is the side lobe gain, and $\omega_s$ is the main lobe beam width.  We set $G_{max}$  as 18 dB \cite{sun2020spatial} and the noise power $N$  in Rx follows a Gaussian distribution with mean of $- 57$ dBm  \cite{sun2017smart} and variance of 1. Due to the large path loss in mmWave communications, the distance between Tx and Rx (i.e., $d$) is generally a short distance. Without loss of generality, we set $d$ as 0.01 kilometers (i.e., 10 meters) in our setup. $G_{min}$ is set to 8 dB. Based on Equation (\ref{rss}), the RSS  under different communication frequencies can be obtained and illustrated in Table \ref{PL}. 

\begin{table}[ht] 
\caption{RSS under different communication frequencies.}
\label{PL}
\centering  

\begin{tabular}{|c|c|c|}  
\hline
Communication frequency (GHz) & Main/side lobe  & RSS  \\ 
\hline
24.25  & Main & $N(0.6103, 1)$ \\
\hline 
24.25  & Side  & $N(0.0610, 1)$\\
\hline 
43.5 &  Main   & $N(0.1897, 1)$\\
\hline 
43.5 &  Side   &  $N(0.0190, 1)$ \\
\hline 
60 &  Main   & $N(0.0997, 1)$\\
\hline 
60 &  Side  & $N(0.0100, 1)$\\
\hline 
\end{tabular}
\end{table}

\begin{figure}[h!]
\centering~
\includegraphics[width=0.65\linewidth]{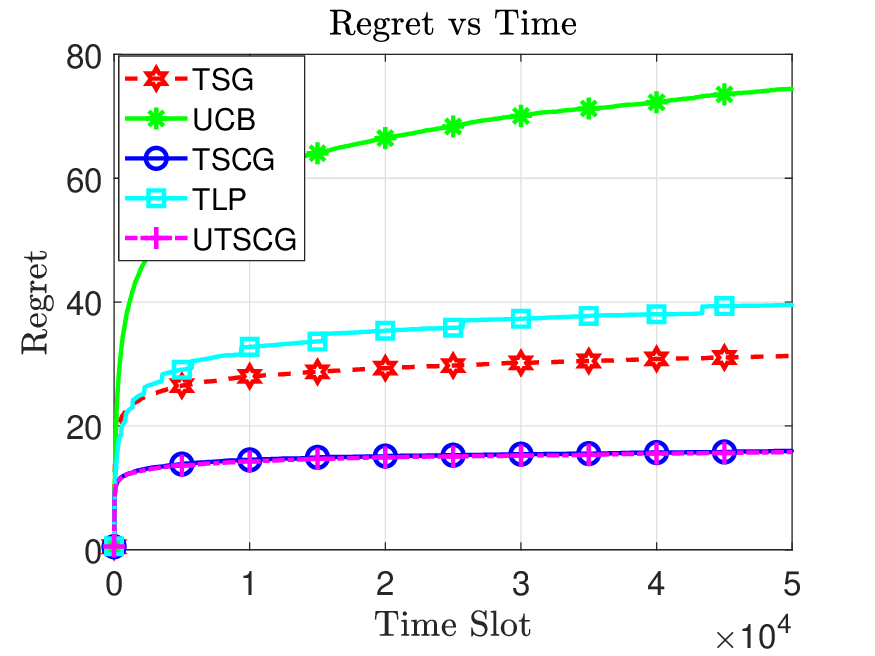}
\caption{Cumulative regret.}
\label{fig:mm_regret}
\end{figure}
\begin{figure}[h!]
    \centering~
    \includegraphics[width=0.65\linewidth]{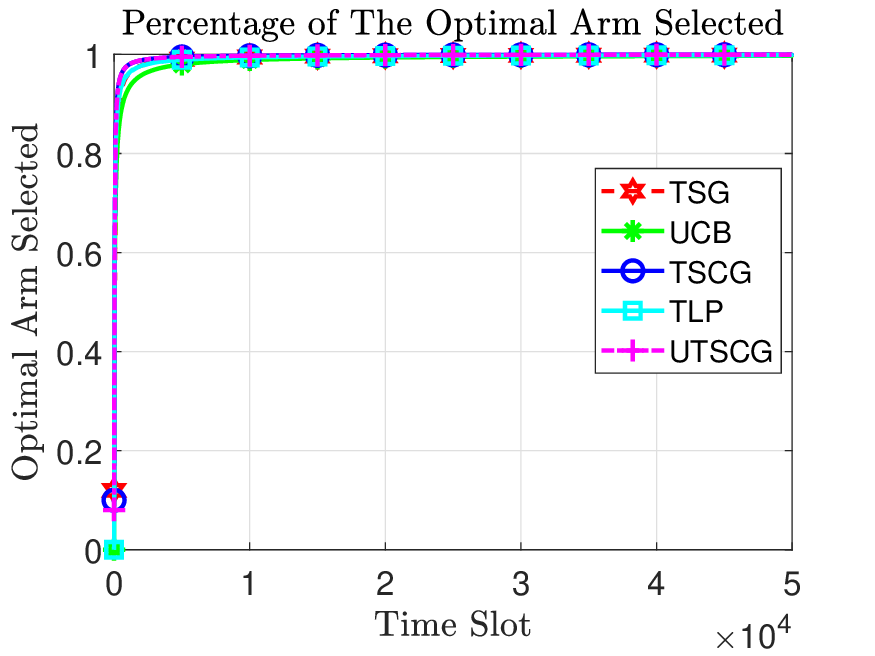}
    \caption{Cumulative selection of the true optimal.}
    \label{fig:mm_rate}
\end{figure}

In Fig.\ref{fig:mm_regret}, points on TSCG and UTSCG are all lower than those of  existing algorithms, which means our proposed algorithms perform better than others, i.e., our modification is a success. Also, incorporating cluster structure indeed lowers the regret significantly. This observation verifies our improvement in terms of the regret bound stated in Theorem 2 and Theorem 3, which mainly arise from the effect of the cluster structure. Fig.\ref{fig:mm_rate} shows that our upgrades are effective: the optimal arm is reached at an earlier time slot than that in other algorithms.  We can see that UTSCG and TSCG share the similar result since each cluster in our setup has only three arms and the number of neighbors for each optimal beam in each cluster is two, so we can not see a clear advantage of Unimodality.

\vspace{-0.3cm}
\subsection{Portfolio Selection}
We prepare a 20-arm system with 4 clusters and there are 5 arms in each cluster. Due to space limitations we leave the details of the arm configuration in supplementary materials. In addition to the number or arms per cluster, another difference w.r.t. the arm setting in the previous case is that we do not restrict clusters to satisfy strong dominance, which means the $\mu$ value of an arm in the non-optimal cluster can be greater than some values of non-optimal arms in the optimal cluster. This is our attempt to test the performance of our algorithms in {a more generalized} arm configuration.
\par

\begin{figure}[h!]
\centering~
\includegraphics[width=0.65\linewidth]{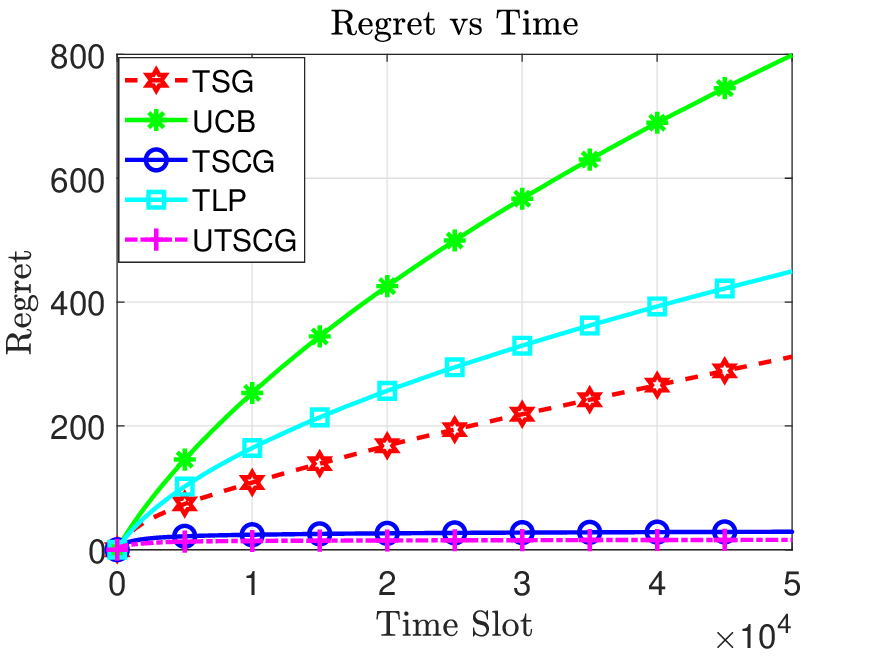}
\caption{Cumulative regret.}
\label{fig:PS_regret}
\end{figure}

\begin{figure}[h!]
    \centering~
    \includegraphics[width=0.65\linewidth]{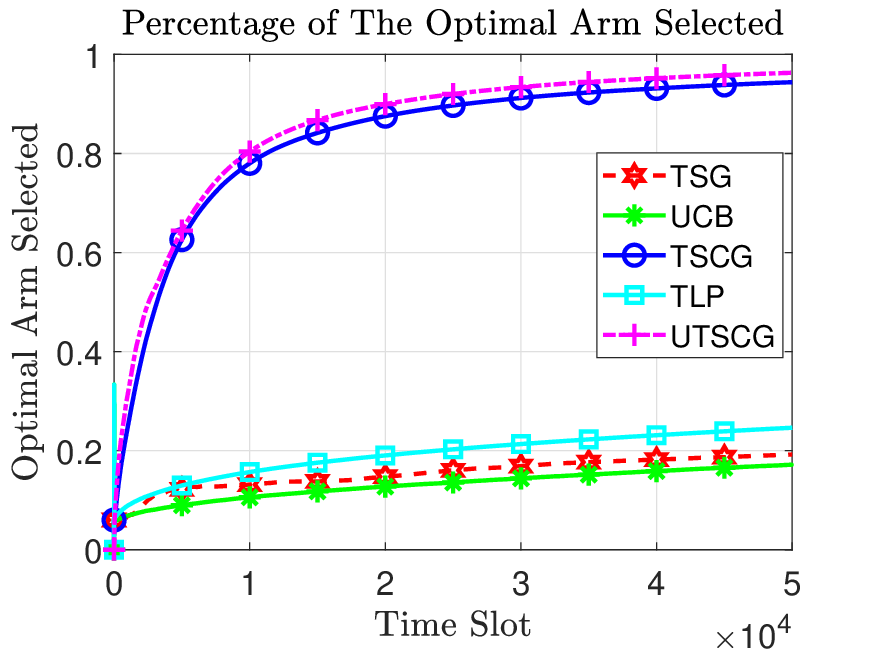}
    \caption{Cumulative selection of the true optimal.}
    \label{fig:PS_rate}
\end{figure}

In Fig.~\ref{fig:PS_regret}, we observe that the vertical gap between our proposed algorithms and the competitors are greatly enlarged compared to results in Fig.~\ref{fig:mm_regret}, which supports our regret analysis. The advantage in regret of UTSCG over TSCG can be seen in the gap between the two algorithms in Fig.~\ref{fig:PS_regret}, however in Fig.~\ref{fig:mm_regret} this discrepancy is not visually explicit.
According to Fig.~\ref{fig:PS_rate}, it takes our proposed algorithms about 25,000 rounds to reach 90\%, while others have no sign of going beyond 30\% in the time window.
\section{Discussions}
\label{sec:discussion}
We show the advantage of our algorithms in regret and the improvement is supported by numerical experiments. The regret in sampling arms can be greatly suppressed by TSCG and UTSCG. In Fig.~\ref{fig:PS_regret}, there is a subtle difference in the cumulative regret of our two proposed algorithms. This overlap can be explained by comparing terms in Eqn.~\ref{eq:proof8} and Eqn.~\ref{eq:proof29}: the upper bound of the cumulative regret in TSCG sums over all arms in the optimal cluster, while in UTSCG only the neighboring arms of the optimal contribute to the overall regret. As a result, the difference is more significant when the optimal cluster contains more arms. Also, since TSCG already outperforms the baseline algorithms significantly, our work can be applied to systems of clustered arms even without Unimodality.
\par
Some of our setup simplifies the real-world  scenario, but this is a good starting point. An immediate and doable upgrade of our work would be generalizing TSCG to other distributions that satisfy the sub-Gaussian property. Another extension is to consider the case in which arms' standard deviations are unknown.

\bibliography{sn-bibliography}
\appendix
\section{Definitions}
\textbf{Fact 1} (Chernoff-Hoeffding bound for sub-Gaussian) Let $X_1...X_n$ be i.i.d  $N(\mu,\sigma^2)$, $\hat{\mu}=\frac{1}{n}\sum_{i=1}^{n}X_i$. Then for all $t\geq 0$,
\begin{eqnarray}
\label{eq:fact2}
&&P(|\hat{\mu}-\mu|>t) \leq 2e^{-\frac{nt^2}{2\sigma^2}} \end{eqnarray}

\textbf{Definition 1} (Quantities $x_i,y_i$). For each arm $i$, we choose two thresholds $x_i$ and $y_i$ such that $\mu_i<x_i<y_i<\mu_{i^*}$. The choice of these thresholds is specific to the algorithm of interest. 

\textbf{Definition 2} (Events $E_i^{\mu}(t),E_i^{\theta}(t)$). 
We define $E_i^{\mu}(t)$ as the event $\hat{\mu}_i(t)\leq x_i$, and  $E_i^{\theta}(t)$ to be the event $\theta_i(t)\leq y_i$.

\section{Proof of Lemma 1}
We develop Lemma 1 because Lemma 3 in \cite{Agrawal2012FurtherThompson} is only suitable for the Beta distribution, and their theoretical analysis for the Gaussian distribution does not provide an extension of Lemma 3.

To derive Lemma 1, we set $x_i=\mu_i+\frac{\Delta_i}{3}$ and $y_i=\mu_{i^*}-\frac{\Delta_i}{3}$ where $\Delta_i=\mu_{i^*}-\mu_i$. Let $\tau_k$ denote the time at which $k$-th trial of arm $i$ happens. Let $\tau_0=0$. Note that $\tau_k>T$ for $k>N_i(T)$. Also, $T\leq \tau_T$. Then, 

\begin{eqnarray}
\label{eq:lemma1_1_repeat}
\sum_{t=1}^{T}P(i(t)=i,\overline{E_i^{\mu}(t)})&&\stackrel{(a)}\leq 1+E[\sum_{k=1}^{T-1}I(\overline{E_i^{\mu}(\tau_k +1)})]\nonumber
\\
&&\stackrel{(b)}\leq 1+\sum_{k=1}^{T-1}2e^{-\frac{k\Delta_i^2}{18\delta^2}}\nonumber
\\
&&=1+\frac{2e^{-\frac{\Delta_i^2}{18\delta^2}}(1-e^{-\frac{(T-1)\Delta_i^2}{18\delta^2}})}{1-e^{-\frac{\Delta_i^2}{18\delta^2}}}\nonumber
\\
&&\leq 1+\frac{2e^{-\frac{\Delta_i^2}{18\delta^2}}}{1-e^{-\frac{\Delta_i^2}{18\delta^2}}}=O(1).  
\end{eqnarray}
Inequality (a) is based on the result from lemma 3 in \cite{Agrawal2012FurtherThompson}. Using Fact 1's result, we obtain inequality (b).
\section{Proof of Theorem 1}
Theorem 1 mainly follows the proof of Theorem 3 in \cite{Agrawal2012FurtherThompson}. The key difference is that we apply our lemma 1 to bound $\sum_{t=1}^{T}P(i(t)=i,\overline{E_i^{\mu}(t)})$, which counts the times when the empirical mean $\hat{\mu}_i(t)$ is far above its expectation. Here, we choose $L_i(T)=\frac{18\log(T\Delta_i^2)}{\Delta_i^2}$ and, by applying Lemma 5 and Lemma 6 from \cite{Agrawal2012FurtherThompson} along with our Lemma 1, we have 
\begin{eqnarray}
\label{eq:Thm1_1}
&&E[N_i(T)]\leq\nonumber
\\
&& e^{11}+5+\frac{18\log(T\Delta_i^2)}{\Delta_i^2}+\frac{1}{\Delta_i^2}+1+\frac{2e^{-\frac{\Delta_i^2}{18\delta^2}}}{1-e^{-\frac{\Delta_i^2}{18\delta^2}}}.    
\end{eqnarray}
Then, the expected regret bound is:
\begin{small}
\begin{eqnarray}
\label{eq:Thm1_2_repeat}
E[R(T)]&&=\sum_{i=1}^{n}\Delta_iE[N_i(T)]\nonumber
\\
&&\leq \sum_{i=1}^{n}\Delta_i\left(e^{11}+5+\frac{18\log(T\Delta_i^2)}{\Delta_i^2}+\frac{1}{\Delta_i^2}\right.\nonumber
\\
&&\left. +1+\frac{2e^{-\frac{\Delta_i^2}{18\delta^2}}}{1-e^{-\frac{\Delta_i^2}{18\delta^2}}}\right)\nonumber
\\
&&=\sum_{i=1}^{n}\Delta_i\left(\frac{18\log(T\Delta_i^2)}{\Delta_i^2}+\frac{1}{\Delta_i^2}+\frac{2e^{-\frac{\Delta_i^2}{18\delta^2}}}{1-e^{-\frac{\Delta_i^2}{18\delta^2}}}\right)\nonumber
\\
&&+O(n)\nonumber
\\
&&=\sum_{i=1}^{n}\left(\frac{18\log(T\Delta_i^2)+1}{\Delta_i}+\frac{2\Delta_ie^{-\frac{\Delta_i^2}{18\delta^2}}}{1-e^{-\frac{\Delta_i^2}{18\delta^2}}}\right)\nonumber
\\
&&+O(n)\nonumber
\\
&&\stackrel{(a)}\leq\sum_{i=1}^{n}\left(\frac{18\log(T\Delta_i^2)}{\Delta_i}\right)+O(n). \nonumber
\\
&&\leq\sum_{i=1}^{n}\left(\frac{18\log(D_{max}T)}{\Delta_i}\right)+O(n).    
\end{eqnarray}\end{small}
Where $D_{max}=\max_{i}\Delta_i^2$. Inequality (a) is based on the fact that $\frac{2\Delta_ie^{-\frac{\Delta_i^2}{18\delta^2}}}{1-e^{-\frac{\Delta_i^2}{18\delta^2}}}$ is a constant, so it can be absorbed by $O(n)$.
\section{Proof of Lemma 2}
\textbf{Step 1: Decomposition}
Assume $C^*$  is the cluster containing the optimal arm  $i^*$. We want to bound $E[N_C]$ for some sub-optimal cluster $C$. Let $\mu_{\underline{C}}$ and $\mu_{\overline{C}}$ be the smallest and largest means in cluster $C$,respectively,and let $\theta_{C}(t)$ be the sample drawn from the belief of TSCG for cluster $C$ at time $t$. We set $x_C=\mu_{\overline{C}}+\frac{\Delta_C}{3}$, $y_C=\mu_{\underline{C^*}}-\frac{\Delta_C}{3}$

If cluster $C$ is played at time $t$, then one of three events needs to happen.
\begin{enumerate}
    \item The event $I(C(t)=C,\hat{\mu}_C(t)\leq x_C, \theta_{C}(t)\leq y_C)$: both the empirical mean and the sampled value are not too far above their expectations.
    \item  The event $I(C(t)=C,\hat{\mu}_C(t)\leq x_C, \theta_{C}(t)> y_C)$: when $\theta_{C}(t)$ is far above its expectation. 
    \item The event $I(C(t)=C,\hat{\mu}_C(t)> x_C)$: the empirical mean is far above its expectation.
\end{enumerate}

Thus, the expected number of pulls, $N_C$, of cluster $C$ can be decomposed as
\begin{eqnarray}
\label{eq:Lemma2_1}
E[N_C(T)]&&\leq \underbrace{\sum_{t=1}^{T}P(C(t)=C,\hat{\mu}_C(t)\leq x_C, \theta_{C}(t)\leq y_C)}_{A}\nonumber
\\
&&+\underbrace{\sum_{t=1}^{T}P(C(t)=C,\hat{\mu}_C(t)\leq x_C, \theta_{C}(t)> y_C)}_{B}\nonumber
\\
&&+\underbrace{\sum_{t=1}^{T}P(C(t)=C,\hat{\mu}_C(t)> x_C)}_{C}
\end{eqnarray}
We define $\preceq$ to denote stochastic domination, i.e. $X\preceq Y$ if and only if $P(X\geq x)\leq P(Y\geq x)$ for all $x$. Then, we bound the three terms as follows:

\textbf{Step 2: Bounding term A}: 
\begin{eqnarray}
\label{eq:Lemma2_2}
&&\sum_{t=1}^{T}P(C(t)=C,\hat{\mu}_C(t)\leq x_C, \theta_{C}(t)\leq y_C)\nonumber
\\
&&\leq \sum_{t=1}^{T}P(C(t)=C,\theta_{C}(t)\leq y_C)\nonumber
\\
&&\stackrel{(a)}\leq \sum_{t=1}^{T}P(C(t)=C,\theta_{\underline{C}}(t)\leq y_C)\nonumber
\\
&&=\sum_{t=1}^{T}P(C(t)=C,\theta_{\underline{C}}(t)\leq y_C,\hat{\mu}_{\underline{C}}(t)\leq x_C)+\nonumber
\\
&&\sum_{t=1}^{T}P(C(t)=C,\theta_{\underline{C}}(t)\leq y_C,\hat{\mu}_{\underline{C}}(t)> x_C)\nonumber
\\
&&\leq \underbrace{\sum_{t=1}^{T}P(C(t)=C,\theta_{\underline{C}}(t)\leq y_C,\hat{\mu}_{\underline{C}}(t)\leq x_C)}_{A_1}+\nonumber
\\
&&\underbrace{\sum_{t=1}^{T}P(C(t)=C,\hat{\mu}_{\underline{C}}(t)> x_C)}_{A_2}\nonumber
\\
&&\stackrel{(b)}\leq O(1),    
\end{eqnarray}
Inequality (a) is based on the fact of stochastic domination. Inequality (b) is based on Lemma 6 from \cite{Agrawal2012FurtherThompson} (for $A_1$)  and Lemma 1 (for $A_2$). 

\textbf{Step 3: Bounding term B}:
\begin{eqnarray}
\label{eq:Lemma2_3}
&&\sum_{t=1}^{T}P(C(t)=C,\hat{\mu}_C(t)\leq x_C, \theta_{C}(t)> y_C)\nonumber
\\
&&\leq \sum_{t=1}^{T}P(C(t)=C,\theta_{C}(t)> y_C)\nonumber
\\
&&\stackrel{(a)}\leq\sum_{t=1}^{T}P(C(t)=C,\theta_{\overline{C}}(t)\leq y_C)\nonumber
\\
&&=\sum_{t=1}^{T}P(C(t)=C,\theta_{\overline{C}}(t)\leq y_C,\hat{\mu}_{\overline{C}}(t)\leq x_C)+\nonumber
\\
&&\sum_{t=1}^{T}P(C(t)=C,\theta_{\overline{C}}(t)\leq y_C,\hat{\mu}_{\overline{C}}(t)> x_C)\nonumber
\\
&&\leq \underbrace{\sum_{t=1}^{T}P(C(t)=C,\theta_{\overline{C}}(t)\leq y_C,\hat{\mu}_{\overline{C}}(t)\leq x_C)}_{B_1}+\nonumber
\\
&&\underbrace{\sum_{t=1}^{T}P(C(t)=C,\hat{\mu}_{\overline{C}}(t)> x_C)}_{B_2}\nonumber
\\
&&\stackrel{(b)}\leq \frac{18\log(T\Delta_{C}'^2)}{\Delta_{C}'^2}+\frac{1}{\Delta_{C}'^2}+O(1), 
\end{eqnarray}
Where $\Delta_{C}'=\mu_{\underline{C^*}}-\mu_{\overline{C}}$. Inequality (a) is based on the fact of stochastic domination. Inequality (b) is based on Lemma 4 from \cite{Agrawal2012FurtherThompson} (for $B_1$) and Lemma 1 (for $B_2$). 

\textbf{Step 4: Bounding term C}:
\begin{eqnarray}
\label{eq:Lemma2_4}
&&\sum_{t=1}^{T}P(C(t)=C,\hat{\mu}_C(t)> x_C)\stackrel{(a)}\leq O(1), 
\end{eqnarray}
Inequality (a) is based on the result of Lemma 1. Combining (\ref{eq:Lemma2_2}), (\ref{eq:Lemma2_3}) and (\ref{eq:Lemma2_4}), we get:
\begin{eqnarray}
\label{eq:Lemma2_5_repeat}
E[N_C(T)]&&\leq \frac{18\log(T\Delta_{C}'^2)}{\Delta_{C}'^2}+O(1)\stackrel{(a)}\leq \frac{18\log(D_{max}T)}{\Delta_{C}'^2}+O(1),
\end{eqnarray}
Where Inequality (a) is based on the fact that $\max_{C'}\Delta_{C}'^2\leq \max_{i}\Delta_i^2=D_{max}$
\section{Proof of Theorem 2}
Similar to the result presentation in \cite{carlsson2021thompson}, we split the regret into the following components:
\begin{eqnarray}
\label{eq:proof1}
&&E[R(T)]=\sum_{i\neq i^*}\Delta_{i}E[\sum_{t=1}^{T}I(i(t)=i)]\nonumber
\\
&&=\sum_{C\neq C^*}\sum_{i\in C}\Delta_{i}E[\sum_{t=1}^{T}I(i(t)=i)]+\sum_{i\in C^*}\Delta_{i}E[\sum_{t=1}^{T}I(i(t)=i)].    
\end{eqnarray}
Here, the first term considers the regret suffered from playing sub-optimal clusters, and the second term represents the regret suffered from playing sub-optimal arms within the optimal cluster. The second term can be bounded by simply applying Theorem 1:
\begin{eqnarray}
\label{eq:proof2}
&&\sum_{i\in C^*}\Delta_{i}E[\sum_{t=1}^{T}I(i(t)=i)]\leq \sum_{i\in C^*}\frac{18\log(T\Delta_i^2)}{\Delta_i}+O\left(|C^*|\right).   \end{eqnarray}
Here $|C^*|$ represents the size of the cluster $C^*$. To bound the first term, consider a sub-optimal cluster $C$ and let $N_C(T)$ denote the number of times we play cluster $C$. Let $i_C^{*}$ be the action with the highest expected reward in $C$. Then for any other $i\in C$, where $i\neq i_C^{*}$, we can bound the number of plays, $N_i(T)$, by applying Theorem 1:
\begin{eqnarray}
\label{eq:proof3}
&&E[N_i(T)]\stackrel{(a)}\leq \frac{18\log(E[N_C(T)]\Delta_i^2)}{\Delta_i^2}+O\left(1\right),   \end{eqnarray}
Inequality (a) is based on the fact that $\log$ function is strictly concave, so $\log(E[x])\geq E[\log (x)]$ (this is Jensen's inequality for concave function). For $i_C^*$, we have:
\begin{eqnarray}
\label{eq:proof4}
&&E[N_{i_C^*}(T)]\leq E[N_C(T)], \end{eqnarray}
From Lemma 2, we know that:
\begin{eqnarray}
\label{eq:proof5}
&&E[N_C(T)]\leq \frac{18\log(T\Delta_{C}'^2)}{\Delta_{C}'^2}, \end{eqnarray}
where inequality (a) is based on the fact that $\log\log (T)\leq \log (T)$ when $T\geq 1$ and we thus get a $\log\log (T)$ dependence on all arms in $C$ except the one with highest expected reward.
\begin{eqnarray}
\label{eq:proof6}
&&E[N_i(T)]\leq\frac{\log\{\frac{18\log(T\Delta_{C}'^2)}{\Delta_{C}'^2}\}}{\Delta_i^2}+O\left(1\right)\nonumber
\\
&&=\frac{\log\log(T\Delta_{C}'^2)}{\Delta_i^2}+\frac{\log\{\frac{18\log (\Delta_{C}'^2)}{\Delta_{C}'^2}\}}{\Delta_i^2}+O\left( 1\right)\nonumber
\\
&&\stackrel{(a)}\leq\frac{\log\log(D_{max}T)}{\Delta_i^2}+ O\left( 1\right)\nonumber
\\
&&E[N_{i_C^*}(T)]\leq E[N_C(T)]\leq \frac{18\log(T\Delta_{C}'^2)}{\Delta_{C}'^2},    
\end{eqnarray}
 Inequality (a) is based on the fact that the second term is constant value, so it can be absorbed by the third term $O(1)$. Therefore, we can bound the regret suffered from sub-optimal clusters:
\begin{eqnarray}
\label{eq:proof7}
&&\sum_{C\neq C^*}\sum_{i\in C}\Delta_{i}E[\sum_{t=1}^{T}I(i(t)=i)]\nonumber
\\
&&\leq\sum_{C\neq C^*}\left\{\sum_{i\in C\ and\ i\neq i_C^*}\left(\frac{\log\log(D_{max}T)}{\Delta_i^2}\Delta_i+ O\left( 1\right)\right)\right.\nonumber
\\
&&\left.+E[N_{i_C^*}(T)]\Delta_{i_C^*}\right\}\nonumber
\\
&&\leq\sum_{C\neq C^*}\left\{\sum_{i\in C\ and\ i\neq i_C^*}\left(\frac{\log\log(D_{max}T)}{\Delta_i}+ O\left( 1\right)\right)\right.\nonumber
\\
&&\left.+\frac{18\log(T\Delta_{C}'^2)}{\Delta_{C}'^2}\Delta_{i_C^*}\right\} \nonumber
\\
&&\leq\sum_{C\neq C^*}\frac{18\log(T\Delta_{C}'^2)}{\Delta_{C}'^2}\Delta_{i_C^*}+o(\log(T))+O\left(1\right)\nonumber
\\
&&\stackrel{(a)}\leq\sum_{C\neq C^*}\frac{18\log(T\Delta_{C}'^2)}{\Delta_{C}'^2}\Delta_{C}+o(\log(T))+O\left(1\right).  
\end{eqnarray}
Inequality (a) is based on the fact that $\Delta_{i_C^*}\leq\Delta_C$. When combining with the bound on regret suffered from sub-optimal clusters, we get:
\begin{eqnarray}
\label{eq:proof8_repeat}
&&E[R(T)]\leq\sum_{C\neq C^*}\frac{18\log(T\Delta_{C}'^2)}{\Delta_{C}'^2}\Delta_{C}+o(\log(T))\nonumber
\\
&&+ \sum_{i\in C^*}\frac{18\log(T\Delta_i^2)}{\Delta_i}+O\left( n\right)\nonumber
\\
&&\leq \sum_{C\neq C^*}\frac{18\log(D_{max}T)}{\Delta_{C}'^2}\Delta_{C}+o(\log(T))+ \nonumber
\\
&&\sum_{i\in C^*}\frac{18\log(D_{max}T)}{\Delta_i}+O\left( n\right).  
\end{eqnarray}
\section{Proof of Theorem 3}
The proof procedure of UTSCG is similar to TSCG. We split the regret as shown in Eq.~\ref{eq:proof1}. The second term in Eq.~\ref{eq:proof1} can be bounded by applying a similar method to the result presentation in Ref. \cite{trinh2020solving}. We split the second term in two sets: those rounds in which the best arm $i^*$ is the leader, and those in which the leader is another arm $i\neq i^*$. Therefore:
\begin{eqnarray}
\label{eq:proof21}
&&\sum_{i\in C^*}\Delta_{i}E[\sum_{t=1}^{T}I(i(t)=i)]\nonumber
\\
&&=\sum_{i\in C^*\ and\ i\neq i^*}\Delta_{i}E[\sum_{t=1}^{T}I(L(t)=i^*\ and\ i(t)=i)]\nonumber
\\
&&+\sum_{i\in C^*\ and\ i\neq i^*}\Delta_{i}E[\sum_{t=1}^{T}I(L(t)\neq i^*\ and\ i(t)=i)],    
\end{eqnarray}
 When considering the first term, the proposed algorithm operates similarly to the TSG algorithm, but with a restriction to the optimal arm and its surrounding neighborhood. The upper bound of the regret in this case aligns with the one presented in Theorem 1.
\begin{eqnarray} 
\label{eq:proof22}
&&E[R_1(T)]\leq  \sum_{i\in C^*\ and\ i\in Neighbor(i^*)}\frac{18\log(T\Delta_i^2)}{\Delta_i}+O\left(1\right),
\end{eqnarray} 
For the second part, we utilize step 3 from Ref.~\cite{yang2023thompson}, which gives us:
\begin{eqnarray} 
\label{eq:proof23}
E[R_2(T)]&&=\sum_{i\in C^*\ and\ i\neq i^*}\Delta_{i}E[\sum_{t=1}^{T}I(L(t)\neq i^*\ and\ i(t)=i)]\nonumber
\\
&&\stackrel{(a)}\leq \sum_{i\in C^*\ and\ i\neq i^*}\left(\frac{12}{\Delta_i}+\frac{16\log(2\Delta_i)+4}{\Delta_i\sqrt{2\pi}}+2\sqrt{2\pi}\right),
\end{eqnarray}
Inequality (a) is based on the step 3 result in \cite{yang2023thompson}. Combining (\ref{eq:proof22}) and (\ref{eq:proof23}), we have:
\begin{eqnarray} 
\label{eq:proof24}
&&\sum_{i\in C^*\ and\ i\neq i^*}\Delta_{i}E[\sum_{t=1}^{T}I(L(t)\neq i^*\ and\ i(t)=i)]\nonumber
\\
&&\leq \sum_{i\in C^*\ and\ i\in Neighbor(i^*)}\frac{18\log(T\Delta_i^2)}{\Delta_i}+O\left(1\right)\nonumber
\\
&&+\sum_{i\in C^*\ and\ i\neq i^*}\left(\frac{12}{\Delta_i}+\frac{16\log(2\Delta_i)+4}{\Delta_i\sqrt{2\pi}}+2\sqrt{2\pi}\right).
\end{eqnarray}
Next, we will bound the first term in Eq.~\ref{eq:proof1}. First, we apply the result of (\ref{eq:proof6}) to obtain the expected time of $E[N_{i_C^*}(T)]$:
\begin{eqnarray} 
\label{eq:proof25}
&&E[N_{i_C^*}(T)]\leq E[N_C(T)]\leq \frac{18\log(T\Delta_{C}'^2)}{\Delta_{C}'^2},
\end{eqnarray}
Then for any other $i\in C$, $i\neq i_C^{*}$, we bound the number of plays, $N_i(T)$.  According to our algorithm procedure, it performs UTSG algorithm under cluster $C$. If $i\in Neighbor(i_C^*)$, we apply the result of Inequality (\ref{eq:proof3}) and Inequality (\ref{eq:proof6}):
\begin{eqnarray}
\label{eq:proof26}
&&E[N_i(T)]\leq \frac{\log\log(D_{max}T)}{\Delta_i^2}+ O\left( 1\right),   \end{eqnarray}
If $i\notin Neighbor(i_C^*)$, we apply the result of step 3 in Ref. \cite{yang2023thompson} and obtain:
\begin{eqnarray}
\label{eq:proof27}
&&E[N_i(T)]\leq \frac{12}{\Delta_i^2}+\frac{16\log(2\Delta_i)+4}{\Delta_i^2\sqrt{2\pi}}+\frac{2\sqrt{2\pi}}{\Delta_i^2}=D_i.   \end{eqnarray}
Therefore, we can bound the regret suffered from sub-optimal clusters:
\begin{eqnarray}
\label{eq:proof28}
&&\sum_{C\neq C^*}\sum_{i\in C}\Delta_{i}E[\sum_{t=1}^{T}I(i(t)=i)]\nonumber
\\
&&=\sum_{C\neq C^*}\left\{ \left(\sum_{i\in C\ and\ 
 i\in Neighbor(i_C^*)}E[N_i(T)]\Delta_i \right.\right.\nonumber
\\
&&\left.\left.+\sum_{i\in C\ and\ 
 i\notin Neighbor(i_C^*)}E[N_i(T)]\Delta_i\right)+E[N_{i_C^*}(T)]\Delta_{i_C^*}\right\}
\nonumber
\\
&&\leq\sum_{C\neq C^*}\left\{\sum_{i\in C\ and\ 
 i\in Neighbor(i_C^*)}\left(\frac{\log\log(D_{max}T)}{\Delta_i^2}+ O\left( 1\right)\right)\right.\nonumber
\\
&&\left.+\sum_{i\in C\ and\ 
 i\notin Neighbor(i_C^*)}D_i\right\}+\sum_{C\neq C^*}E[N_{i_C^*}(T)]\Delta_{i_C^*}\nonumber
\\
&&\leq\sum_{C\neq C^*}\left\{\sum_{i\in C \ and \ i\neq i_C^*}\left(\sum_{i\in Neighbor(i_C^*)}\left(\frac{\log\log(D_{max}T)}{\Delta_i^2}\right)\right)\right\}\nonumber
\\
&&+ \sum_{C\neq C^*}\left\{\sum_{i\in C \ and \ i\neq i_C^*}\left(\sum_{i\in Neighbor(i_C^*)}O\left( 1\right)\right)\right\}\nonumber
\\
&&+\sum_{C\neq C^*}\left\{\frac{18\log(T\Delta_{C}'^2)}{\Delta_{C}'^2}\Delta_{i_C^*}\right\}+F'\nonumber
\\
&&\leq\sum_{C\neq C^*}\frac{18\log(T\Delta_{C}'^2)}{\Delta_{C}'^2}\Delta_{i_C^*}+o(\log(T))+O\left(n\right)+F'\nonumber
\\
&&\leq\sum_{C\neq C^*}\frac{18\log(T\Delta_{C}'^2)}{\Delta_{C}'^2}\Delta_{C}+o(\log(T))+O\left(n\right)+F',
\end{eqnarray}

where: 
\begin{eqnarray}
&&F'=\sum_{i\in C\ and\ 
 i\notin Neighbor(i_C^*)}\Delta_iE[N_i(T)]\nonumber
\\
&&=\sum_{i\in C\ and\ 
 i\notin Neighbor(i_C^*)}\left(\frac{12}{\Delta_i}+\frac{16\log(2\Delta_i)+4}{\Delta_i\sqrt{2\pi}}+2\sqrt{2\pi}\right). \nonumber   
\end{eqnarray}
 Combining (\ref{eq:proof24}) and (\ref{eq:proof28}), we have:
\begin{eqnarray}
\label{eq:proof29_repeat}
&&E[R(T)]\leq\sum_{C\neq C^*}\frac{18\log(D_{max}T)}{\Delta_{C}'^2}\Delta_{C}+o(\log(T))\nonumber
\\
&&+ \sum_{i\in Neighbor(i^*)}\frac{18\log(D_{max}T)}{\Delta_i}+O\left(n\right)+F.  
\end{eqnarray}
Where $F=\sum_{i\neq i^*}\left(\frac{12}{\Delta_i}+\frac{16\log(2\Delta_i)+4}{\Delta_i\sqrt{2\pi}}+2\sqrt{2\pi}\right)$
\section{Portfolio Selection Arm Configurations}
We set the standard deviation $\sigma = 1$ across all arms. The mean parameter for the reward distribution for all arms are in the table below.
\begin{table}[t]
    \centering
    \begin{tabular}{ccc}
    \hline
       arm\#  & cluster & mean\\
       \hline
       1&1  &0.060\\
       2&1  &0.063\\
       3&1  &0.070\\
       4&1  &0.067\\
       5&1  &0.065\\
       6&2  &0.036\\
       7&2  &0.042\\
       8&2  &0.044\\
       9&2  &0.040\\
       10&2&0.038\\
       11&3&-0.02\\
       12&3&0.00\\
       13&3&0.02\\
       14&3&0.04\\
       15&3&0.06\\
       16&4&-0.028\\
       17&4&-0.026\\
       18&4&-0.022\\
       19&4&-0.024\\
       20&4&-0.030\\
       \hline 
    \end{tabular}
    \caption{Arm configuration of the portfolio selection experiment.}
    \label{tab:ps}
\end{table}
\end{document}